\documentclass{article}

\usepackage[english]{babel}

\usepackage[letterpaper,top=2cm,bottom=2cm,left=3cm,right=3cm,marginparwidth=1.5cm]{geometry}
\usepackage{subcaption}
\usepackage{graphicx,booktabs,tabulary,tabularx}

\usepackage{amsmath}
\usepackage{graphicx}
\usepackage[colorlinks=true, allcolors=blue]{hyperref}
\usepackage{makecell}
\usepackage{comment}
\title{Leveraging Self-Training and Variational Autoencoder for Agitation Detection in People with Dementia Using Wearable Sensors}
 
{\author{Abeer Badawi $^1$, Somayya Elmoghazy$^1$, Samira Choudhury$^2$ $^3$, Khalid Elgazzar$^1$,\\ and Amer Burhan $^2$ $^3$\\
			$^1$ Computer Engineering, Ontario Tech University, Oshawa, ON, Canada\\
             $^2$ Ontario Shores Centre for Mental Health Sciences, Whitby, ON, Canada \\  $^3$ Temerty Faculty of Medicine, University of Toronto, Toronto, ON, Canada}}

\begin{document}
\maketitle

\begin{abstract}
Dementia is a neurodegenerative disorder that has been growing among elder people over the past decades. This growth profoundly impacts the quality of life for patients and caregivers due to the symptoms arising from it. Agitation and aggression (AA) are some of the symptoms of people with severe dementia (PwD) in long-term care or hospitals. AA not only causes discomfort but also puts the patients or others at potential risk. Existing monitoring solutions utilizing different wearable sensors integrated with Artificial Intelligence (AI) offer a way to detect AA early enough for timely and adequate medical intervention. However, most studies are limited by the availability of accurately labeled datasets, which significantly affects the efficacy of such solutions in real-world scenarios. This study presents a novel comprehensive approach to detect AA in PwD using physiological data from the Empatica E4 wristbands. The research creates a diverse dataset, consisting of three distinct datasets gathered from 14 participants across multiple hospitals in Canada. These datasets have not been extensively explored due to their limited labeling. We propose a novel approach employing self-training and a variational autoencoder (VAE) to detect AA in PwD effectively. The proposed approach aims to learn the representation of the features extracted using the VAE and then uses a semi-supervised block to generate labels, classify events, and detect AA. We demonstrate that combining Self-Training and Variational Autoencoder mechanism significantly improves model performance in classifying AA in PwD. Among the tested techniques, the XGBoost classifier achieved the highest accuracy of 90.16\%. By effectively addressing the challenge of limited labeled data, the proposed system not only learns new labels but also proves its superiority in detecting AA.

%\textcolor{red}{The proposed system aims to learn the representation of the features extracted using a Variational Autoencoder (VAE), then uses a semi-supervised block to classify and detect AA. The proposed system introduces two approaches for AA classification in PwD. In the first approach, supervised and semi-supervised models are trained directly on features extracted from the raw data. The second approach employs VAE for feature refinement, demonstrating the ability of autoencoders to select optimal features and reduce dimensionality. Results showcase the effectiveness of VAE and semi-supervised approaches in classifying AA in PwD with the XGBoost classifier outperforming other techniques with 90.16\% accuracy. By addressing the issue of the scarce labeled data, we show the capability of the proposed system to learn new labels and its proven superiority in AA detection. This allows for the combination of labeled and unlabeled data to build more practical and applicable real-world solutions. Comparisons with prior studies underscore the robustness of the proposed system, achieving a remarkable 99.6\% AUC ROC. The research contributes to the evolving landscape of dementia-related activity classification, providing insights into effective methodologies and their implications for real-world applications.}

\end{abstract}


\section{Introduction}
Dementia is a term that describes a collection of neurodegenerative symptoms that cause a severe cognitive decline that significantly impacts individuals' abilities and quality of life~\cite{WHO2023,gale2018dementia}. Dementia is caused by a range of neurologic, neuropsychiatric, and medical conditions, with Alzheimer's disease and vascular dementia being the most prevalent~\cite{bansal2014dementia}. The progression of the syndrome leads to increasing dependence and a significant burden on patients, caregivers, and society~\cite{kester2009dementia}. The symptoms include cognitive decline, emotional and behavioral changes, and loss of independence. Also, the non-cognitive neuropsychiatric symptoms (NPS) are common in people with dementia (PwD)~\cite{Lyketsos2002,burhan2023s13}. Some examples of NPS are agitation, aggression, apathy, sleep, and appetite disturbances. In severe cases, agitation and aggression (AA) pose a frequent source of distress for the patients and those around them~\cite{Ballard2013,khan2018detecting}. These behaviors are often linked to cognitive impairment, which manifests in various forms such as cursing, hitting, or pacing~\cite{cohen1990dementia}. Although a cure remains elusive, monitoring systems play a crucial role in supporting PwD and people around them to prevent severe aggressive behavior.

Artificial Intelligence (AI) algorithms can be integrated with different wearable sensors to develop predictive algorithms to detect changes in behavior and physical symptoms~\cite{bharucha2009intelligent}. These algorithms can be employed in monitoring systems to assist PwD, detect early signs of AA, and notify caregivers in a timely manner. This allows for quick intervention and prevents the escalation of such behaviors~\cite{khan2018detecting, fabrizio2021artificial}. Integrating wearable sensors into daily living environments has been a groundbreaking development in monitoring PwD~\cite{rezvani2021semi}. Devices like smart wristbands and watches, rings, and patches, collect physiological markers of the patients such as heart rate, skin temperature, and movement data from accelerometers~\cite{Sato2019AutomaticFE,Chikhaoui2017TowardsAF}. 

AI algorithms analyze the collected data to identify the complex behaviors of PwD and detect any early signs of abnormal patterns. However, the development of clinically useful models is impeded by the necessity for extensive labeled data. The collection of such data poses challenges since it requires extensive manual observations that are both time and resource-consuming~\cite{rezvani2021semi}. Consequently, adopting a semi-supervised learning strategy provides a viable solution to these challenges by integrating limited labeled data with a larger pool of unlabeled data~\cite{abdallah2018activity}. This method proves particularly beneficial in scenarios where collecting extensive labeled data is not feasible. It is especially advantageous when dealing with data collected from dementia cases, where, despite the generation of vast amounts of data through continuous monitoring, the manual labeling process constrains the utility of datasets and the effectiveness of supervised learning algorithms.

On the other hand, autoencoders represent a form of unsupervised learning that is highly proficient in feature extraction. They achieve this by learning the representation of input data through compression into a lower-dimensional form and subsequent reconstruction to the original format. This technique aids in extracting and selecting the most relevant features from the data, proving effective in analyzing data from wearable sensors~\cite{Sato2019AutomaticFE, Chikhaoui2017TowardsAF}. These features contribute to the accuracy of classification models to enhance the efficiency of detecting AA and behavioral changes~\cite{Li2020ExtractionAI}. Variational autoencoders (VAEs) exemplify a type of autoencoder capable of learning complex data distributions. The VAE structure consists of both an encoder and a decoder to understand the features of the input data and utilize them for reconstruction. VAEs specifically employ probabilistic encoders and decoders, making them efficient in managing uncertainty in data by learning a robust representation and generating new data samples resembling the input data. These characteristics render VAEs particularly valuable in applications where comprehending complex data is essential~\cite{kingma2013auto}.

In this study, we present a novel Self-training model that uses a VAE to detect AA in PwD using data from wearable sensors. The uniqueness of our approach lies in specialized training on a unique and distinctive not fully labeled dataset~\cite{NCT04516057, NCT03672201}. The architecture of the system is composed of a VAE for feature extraction and selection, followed by a Self-training mechanism. The VAE effectively discerns optimal features from the sensor data to represent behavioral patterns. The performance of the proposed approach is compared to: (1) a baseline model using supervised learning on features extracted from raw data, (2) a supervised learning model enhanced by a VAE for improved feature selection, and (3) a semi-supervised model trained directly on raw data features extracted from the datasets. Three different models are evaluated in both supervised and semi-supervised scenarios.

Our results show that the combination of VAE with semi-supervised learning for AA detection in PwD surpasses the baseline model. These findings not only highlight the potential of merging VAE with semi-supervised learning in AA detection but also emphasize the ability of the system to learn from both labeled and unlabeled data. To the best of our knowledge, this is the first study incorporating autoencoders for feature extraction and semi-supervised learning for AA detection in PwD from wearable sensors. The proposed system opens up new possibilities for continuous monitoring and timely intervention in the realm of dementia by offering essential improvements in caregiving quality. 

The remainder of the paper is organized as follows. Section \ref{s:rw} presents a brief overview of recent research related to the proposed work. Section \ref{s:methods} describes the study design and the proposed system. The system results and discussion are shown in Section \ref{s:numerical}. Lastly, Section \ref{s:conclusion} provides concluding remarks and outlines future directions.

\section{Related Work} \label{s:rw}
Dementia, marked by cognitive decline affecting daily life, often leads to AA in patients~\cite{gale2018dementia,bansal2014dementia, kester2009dementia}. These symptoms, which are linked to factors like mental health history and memory issues, emphasize the need for effective management strategies in dementia care. Wearable sensors have shown promise in detecting and monitoring AA in PwD. A series of studies have demonstrated the potential of wearable sensors in detecting AA in PwD. Spasojevic et al.~\cite{spasojevic2021pilot} found that data from wearable sensors can improve the accuracy of AA detection using personalized models. Nesbitt et al.~\cite{nesbitt201915} found that increases in limb movements and heart rate, as measured by smartwatch technology, correlated well with observations of AA. Khan et al.~\cite{khan2019agitation} further supported this by demonstrating the effectiveness of combining multi-modal sensor data in detecting AA. Furthermore, autoencoders have been shown to outperform traditional feature extraction techniques in the context of wearable sensor data analysis~\cite{zhang2022deep}. Recent studies have demonstrated the potential of autoencoders in extracting features from wearable sensor data. Boning Li et al.~\cite{Li2020ExtractionAI} used a Locally Connected Long Short-Term Memory Denoising AutoEncoder (LC-LSTM-DAE) to predict mood, health, and stress with high precision. Chikhaoui et al.~\cite{chikhaoui2017towards} proposed a deep learning approach for activity recognition, using matrix factorization for dimensionality reduction and a stacked auto-encoder for feature extraction. Wang et al.~\cite{wang2016recognition} introduced a continuous autoencoder (CAE) for human activity recognition, achieving a 99.3\% correct differentiation rate. These studies collectively highlight the effectiveness of autoencoders in extracting features from wearable sensor data for various applications.

On the other hand, semi-supervised learning has gained notable attention in recent years due to its potential to enhance model performance in the presence of limited labeled data~\cite{zhu2022introduction}. This approach enhances model performance by utilizing the information in unlabeled data. Various models have been proposed to enhance semi-supervised learning, including self-training, mixture models, co-training, multiview learning, and graph-based methods~\cite{zhu2022introduction}. Hassanzadeh et al.~\cite{hassanzadeh2018clinical} found that leveraging unlabeled data can significantly reduce the dependency on labeled data, explicitly noting the effectiveness of convolutional neural networks in this context. Many techniques, such as contemporary models, are used to explore semi-supervised learning, which often incorporate unlabeled data into their objective or optimization processes. The Mean Teacher model is an example involving a dual system of teacher and student models~\cite{tarvainen2017mean}. In this setup, the student model updates with the latest weights, while the teacher model maintains an average of these weights. Pseudo-labeling is another technique that can be applied during pre-processing or training. This involves a classifier generating false labels for unlabeled data, then using these labels to train the final model~\cite{triguero2015selflabeled}. This can also form a part of self-training strategies, where the most confident pseudo-labels from a previous iteration are used in the next. However, determining the most reliable confidence measure for this process poses a challenge. Self-supervised representation learning, a type of pseudo-labeling, involves transforming unlabeled data into latent representations, then training a classifier to identify these transformations, aiming to learn generalizable features from the unlabeled data~\cite{sarkar2020selfsupervised}.

In healthcare, semi-supervised learning and autoencoders have diverse applications, including behavior detection, which can be particularly beneficial for PwD. In the broader scope, self-supervised representation learning methods have been utilized for emotion recognition using physiological signals~\cite{quispe2021applying,doe2021semi}. These studies highlight the potential of semi-supervised learning in addressing the labeling data problem in hospital settings. In the human activity recognition (HAR) context, Hinkle et al.~\cite{hinkle2021endtoend} developed a semi-supervised methodology using a wristband. Another paper by Han Yu et al.~\cite{yu2022semi} explores semi-supervised learning methods for stress detection using wearable sensor data. Focusing on dementia care, HekmatiAthar et al.~\cite{hekmatiathar2022data} explored deep learning, particularly LSTM models, for forecasting AA in dementia patients. This method addresses the class imbalance in datasets and aims to predict AA up to 30 minutes in advance.  Roonak Rezvani et al.~\cite{rezvani2021semi} proposed a semi-supervised model, combining self-supervised learning with Bayesian ensemble classification. This model is tailored to analyze and predict AA using in-home monitoring data collected from sensors in the homes of 96 patients. Despite these advancements, the challenge of effectively utilizing the extensive unlabeled data from wearable devices worn by PwD remains.  

To address the challenges in previous works, this research introduces a novel self-training model using a VAE to detect AA in PwD from wearable sensor data. Our model employs a VAE for feature learning and extraction, followed by a self-learning semi-supervised component that predicts pseudo-labels for the unlabeled data and then uses the information for AA classification. Our contributions lie in utilizing three datasets collected from multiple hospital environments of dementia patients wearing wearable wristbands. We examine the efficacy of the VAE in feature extraction compared to different methods. We also incorporate a Self-training block where we test and compare three classification models: Random Forest, Extra Trees, and XGBoost. We prove that the VAE improves the overall performance of the model when used in integration with the self-learning mechanism to aid the classification of the collected digital biomarkers.

\section{Research Methodology}\label{s:methods}

\begin{figure}[h!]
 \centering
  %\vspace*{-.45in}
 \subfloat{\includegraphics[width=14cm]{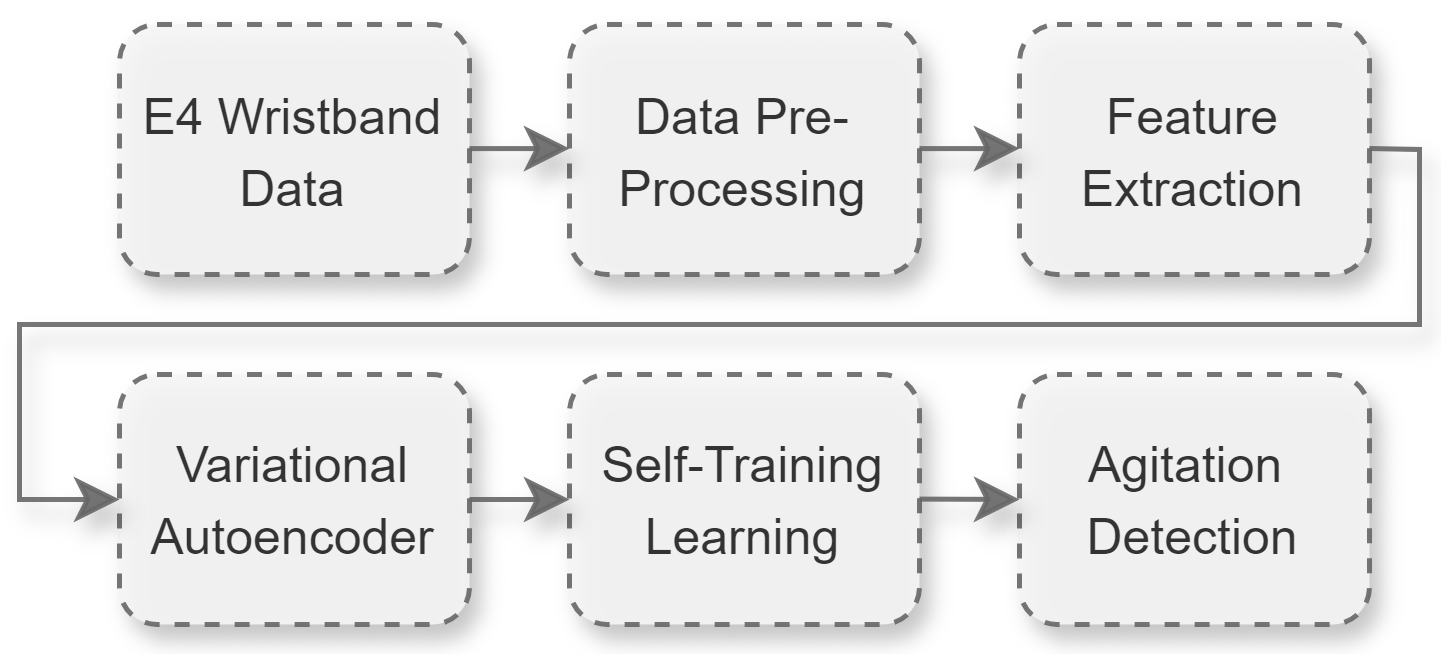}}
 \caption{A Block Diagram of the proposed System Architecture.}
  \label{fig:sysetmall}
\end{figure}

This study introduces a robust methodology for detecting AA in PwD using wearable sensor data collected from the Empatica E4 wristband. The dataset was obtained from three Canadian hospital studies, including the NAB-IT and StaN clinical trials, comprising physiological data such as heart rate, skin conductance, and motion metrics from 14 participants over 48–72 hours of monitoring. First, we present the dataset description in detail, including the number of participants, data collection duration, and labeling of agitation events. Second, we present the proposed system architecture components to classify AA in PwD. We start with the raw wristband data input and the pre-processing steps. Then, we explain the feature extraction techniques used to extract features from the data.
Next, we propose Variational Autoencoders (VAE), which are built on the extracted features to represent the features. We then present the Self-training to classify agitation episodes with labeled and unlabeled data. Finally, we evaluate the performance by comparing the model with the baseline supervised models used to classify the fully labeled dataset and explain the balanced accuracy metrics used to evaluate the performance. Each of these stages is discussed in detail in the following subsections. Figure \ref{fig:sysetmall} shows the overall proposed system architecture.

\subsection{Dataset Description}
%The data was collected from two male participants and one female aged between 80 and 85 and diagnosed with dementia, Alzheimer's, or mixed type. Stringent data confidentiality measures were implemented to safeguard privacy, with securely stored information accessible exclusively to the research team.
The first dataset is a result of a pilot study we conducted at the Ontario Shores Center for Mental Health Sciences. The NAB-IT study, aimed at treating AA in individuals with dementia, is an ongoing randomized, double-blind, placebo-controlled clinical trial recruiting participants nationwide. The StaN study, a randomized, controlled, clinical trial from seven sites across Canada, received approval from institutional review boards at all sites. The participants or their substitute decision-makers provided informed consent before any study procedures. Participants in both trials had their physiological data collected using the Empatica E4 wristband. The wristbands were placed on the non-dominant arm by a study team member during participants' psychometric assessments, and the physiological data stored on the E4 wristband's memory card was later synchronized with clinical records. 

A total of 14 participants were included in this work from distinct studies, with nine male and five female. The data from Ontario Shores Hospital was collected over six weeks. The recruitment process included additional criteria such as the requirement of the patients to have a moderately and severe major neurocognitive disorder defined by a Mini‐Mental State Examination (MMSE) score~\cite{american2013diagnostic}. Other requirements included the ability to ambulate independently, without the assistance of another person, with or without a walking aid. Moreover, participants needed to meet the AA criteria defined by the Agitation Definition Working Group from the International Psychogeriatric Association~\cite{Cummings2015AgitationCognitiveDisorders}. 

The participants wore the E4 device on three occasions, each spanning 48 to 72 hours. The initial vital signs using the E4 wristband were used as a baseline measurement. The clinical staff closely monitored AA in participants during these periods and noted any AA episodes, including the start and end times of these events. The E4 data, which includes the physiological parameters of the participants, was cross-referenced with observational notes for validation. We recorded the data of three participants: for Participant \#1, we recorded 72.13, 70.38, and 50.41 hours across the three sessions; Participant \#2 recorded 47.22, 72.50, and 58.56 hours; and Participant \#3 recorded 24 hours for each of the three sessions. For privacy concerns, all information gathered is kept private and anonymous to all participants.

For the NAB-IT study, the data was recorded over the first eight weeks, with participants wearing the wristband for 48 to 72 hours during each of the three periods. The inclusion criteria for the NAB-IT study included being 55 years of age or older, having a DSM-5 diagnosis of Alzheimer's or mixed dementia, and possessing a Mini-Mental State Examination (MMSE) score of 24 or less~\cite{kurlowicz1999mini}. For the StaN study, the inclusion criteria included being 50 years of age or older, having a clinical diagnosis of Alzheimer's or mixed dementia according to DSM-5, and exhibiting clinically significant AA as per the IPA provisional consensus clinical and research definition~\cite{cummings2015agitation}. The data was recorded in three time periods during the 12-week study, with participants wearing the wristband for 48 to 72 hours at three key points: at the start, midpoint, and conclusion of the study. We then examined documented episodes of AA recorded in nurses' shift notes and Dementia Observation System records (DOS). These data were assessed and labeled as episodes of AA with the exact start and end times. A total of five participants were included in the NAB-IT study, and six participants in StaN met the inclusion criteria and are included in the dataset.

In the collected dataset, five participants were fully labeled, and nine participants were unlabeled. We cleaned the dataset, removed all the missing data, and marked all the labeled data from the wristband data and nurses' notes. We used the confirmed AA events by nurses that include a start and end time to have precise labels for AA events. Table \ref{tab:table1} shows a summary of the dataset. It is composed of a total of 18804 minutes labeled normal, 1475 minutes labeled AA, and 37463 minutes unlabeled data. A total of 57742 minutes is included in our dataset, which are used as basis for the preprocessing steps, shown in the next subsection, to investigate our data. 

% \usepackage{graphicx}
\begin{table}[]
\caption{Overview of the Collected Data.}
\centering
\label{tab:table1}
\resizebox{.9\textwidth}{!}{%
\begin{tabular}{ll}
\hline
Study Place                                         & Multiple hospitals across Canada   \\ 
Number of Participants                        & 14 participants  \\ 
Gender Distribution                           & 9 Male, 5 Female \\ 
Total Normal Labeled Data (minutes)           & 18804            \\ 
Total AA Labeled Data (minutes)        & 1475             \\ 
Total Unlabeled Data (minutes)                & 37463            \\ 
The Maximum Labeled AA Event (minutes) & 217              \\ 
The Minimum Labeled AA Event (minutes) & 2                \\ 
Total Number of Data Collected (minutes)      & 57742            \\ \hline
\end{tabular}%
}
\end{table}

\subsection{Proposed System Architecture}
The proposed system architecture contains multiple blocks to classify AA in PwD. We start with the raw wristband data input and the pre-processing steps. Then, we explain the feature extraction techniques used to extract features from the data. Next, we propose Variational Autoencoders (VAE), which are built on the extracted features to represent the features. We then present the Self-training to classify agitation episodes with labeled and unlabeled data. Finally, we evaluate the performance by comparing the model with the baseline supervised models used to classify the fully labeled dataset and explain the balanced accuracy metrics used to evaluate the performance. Each of these stages is discussed in detail in the following subsections. 

\subsubsection{Data Pre-processing and Feature Extraction}
We selected the Empatica E4 wristband in this study due to its lightweight, portable nature, and high precision compared to alternative devices~\cite{empactica}. The smartwatch is equipped with sensors to monitor vital physiological parameters, including heart rate (measured by a Photoplethysmography sensor at 64 Hz), accelerometer (captured by a 3-axis accelerometer at 32 Hz), skin temperature, and electrical properties of the skin (measured by an Electrodermal Activity sensor at 4 Hz). These physiological signals are collected and can be wirelessly transmitted to a smartphone via Bluetooth for real-time analysis. The recorded data is subsequently uploaded to the Empatica website and can be downloaded from the cloud. The pre-processing steps used in this work have proved their efficiency and ability to classify normal and AA events in PwD in our two previous works~\cite{10288764,10371835}. 

Before data analysis, it is crucial to pre-process the data to ensure reliability. We use the Flirt toolkit~\cite{foll2021flirt}, an up-to-date open-source Python library employed for data processing and feature extraction. Flirt is specifically designed for wearable devices, offering various techniques and a wide range of features to extract necessary information. The primary objective is to perform data cleaning and feature extraction for each attribute collected by the E4 device using state-of-the-art techniques. We start with reading the E4 device data, converting it to the correct format with timestamps, and utilizing the Flirt function for various pre-processing steps, including noise removal and sliding windows. Then, we resample the signals to align with the minimum sampling rate of 4 Hz.  The Flirt function calculates inter-beat intervals (IBI) and applies a low-pass filter for the heart rate. It also uses a low-pass filter with a 10 Hz cut-off frequency for the accelerometer signal. The EDA signal undergoes artifact removal, noise filtering, and decomposition to generate phasic and tonic components. 

\begin{figure}[h!]
 \centering
  %\vspace*{-.45in}
 \subfloat{\includegraphics[width=12cm]{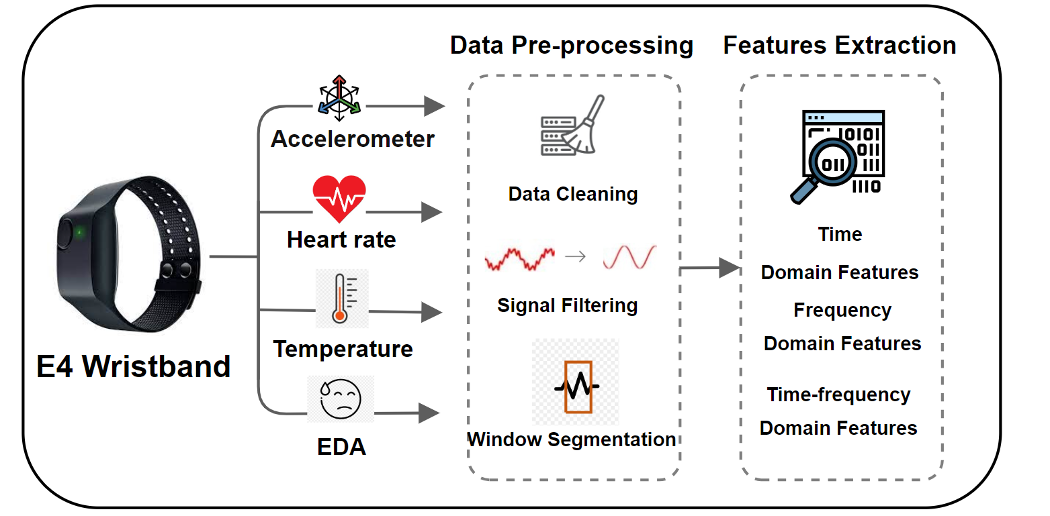}}
 \caption{The Data Pre-processing and Feature Extraction Workflow Derived from Empatica E4 Wristband Signals.}
  \label{fig:features}
\end{figure}

A sliding-window approach with a window size of 1 minute and no overlap is then applied to extract features from the datasets. A feature vector, encompassing the time domain, frequency domain, and statistical features, totaling 198 features, is computed from the four signals. From the 198 features, 88 were from the accelerometer signal, 44 were from the BVP signal, 44 from the EDA signal, and 22 were from the temperature signal. These features cover various domains, including time, frequency, and time-frequency. Out of the total features extracted, only 182 were used in this study. This is because some features yielded inaccurate results, such as NaN or infinity values. The selected features comprehensively represent the physiological markers of the wearable device for subsequent analysis. Figure \ref{fig:features} shows the features engineering steps discussed in this section.

\subsubsection{Variational Autoencoder}
In this work, we use the VAE to evaluate the ability of autoencoders to select the optimal set of features and reduce the dimensions in the AA classification problem. The VAE marks an intriguing advancement in the field of machine learning, bringing together generative modeling and inference. The fundamental concepts of VAE involve the creation of a probabilistic model for data and a VAE model for latent variables. Rooted in Bayesian inference, VAE aims to represent the underlying probability distribution of data, enabling the generation of new data samples from this distribution~\cite{kingma2013auto}. Our proposed system consists of a VAE model incorporating an encoder, sampling function, and a decoder. The system extracts the encoder features and selects the most important features through the model. The encoder part of the VAE is responsible for mapping the input data to the mean (\(z_{\text{mean}}\)) and log variance (\(z_{\text{log\_var}}\)) of the latent space. The input data is first transformed through a dense neural network layer and then passed through a ReLU activation function, as shown in the equation~\ref{eq:h} below. The output is then used to calculate both the mean and the log variance as shown in equations~\ref{eq:z_mean} and~\ref{eq:z_log_var}, providing the necessary parameters for the VAE's probabilistic modeling.

\begin{align}
h &= \text{ReLU}(\text{Dense}(\text{inputs}))  \label{eq:h} \\
z_{\text{mean}} &= \text{Dense}(h) \cite{kingma2019introduction}  \label{eq:z_mean} \\
z_{\text{log\_var}} &= \text{Dense}(h) \cite{kingma2019introduction}  \label{eq:z_log_var}
\end{align}

The input layer has \(input\_dim \) neurons. Dense Layer 1 processes the input through a fully connected layer with 256 neurons, followed by a ReLU activation function. Dense Layers 2 and 3 further reduce dimensionality with 128 and 100 neurons and a ReLU activation. We define a function `sampling` that takes \(z_{\text{mean}}\) and \(z_{\text{log\_var}}\) as arguments and generates a random sample \(z\) from the distribution represented by these parameters as shown in~\ref{z}. The sample is obtained using the reparameterization trick. 
\
\begin{align}
&\ z = z_{\text{mean}} + \exp\left(\frac{1}{2} \times z_{\text{log\_var}}\right) \times \epsilon \cite{doersch2016tutorial} \label{z}
\\ 
&\text{and } \epsilon \sim \mathcal{N}(0, 1) \notag
\end{align}
\
where:\(z\) is the sampled point in the latent space,\(z_{\text{mean}}\) is the mean vector,\(z_{\text{log\_var}}\) is the log variance vector, and \(\epsilon\) is a random sample from a standard normal distribution.\\

The Decoder reconstructs the input data from the samples in the latent space, shown in~\ref{eq:h_decoded}. It mirrors the encoder architecture in reverse order. For Dense Layer 1, it processes the sampled \( z \) with 100 neurons and a ReLU activation. Dense Layers 2 and 3 progressively increase the dimensionality by 128 and 256. The output layer reconstructs the input with a sigmoid activation function as shown in~\ref{eq:x_decoded}. 
   \
   \begin{align}
   h_{\text{decoded}} &= \text{ReLU}(\text{Dense}(z)) \label{eq:h_decoded} \\
   x_{\text{decoded\_mean}} &= \text{Sigmoid}(\text{Dense}(h_{\text{decoded}}))  \label{eq:x_decoded} 
   \end{align}
   \
We use a custom loss function to evaluate the VAE architecture. We start with a reconstruction Loss, which is calculated using binary cross-entropy for each feature. This term is scaled by the total number of features. Then, we apply the Kullback-Leibler Divergence \(kl\_loss\) which encourages the latent variables to follow a specific distribution (typically a multivariate normal distribution). The formula for the KL divergence is: 
 %%  \[ \text{{kl\_loss}} = -0.5 \times \sum_{i=1}^{\text{{latent\_dim}}} (1 + \text{{z\_log\_var}} - \text{{z\_mean}}^2 - e^{\text{{z\_log\_var}}}) \]

\begin{equation}
\text{{kl\_loss}} = -0.5 \times \sum_{i=1}^{\text{{latent\_dim}}} (1 + \text{{z\_log\_var}} - \text{{z\_mean}}^2 - e^{\text{{z\_log\_var}}}) \cite{kingma2013auto} \label{kl_loss}
\end{equation}

where \(latent\_dim\) refers to the dimensionality of the latent space. To calculate the total VAE Loss \(vae\_loss\), we calculate the sum of the reconstruction loss \(xent\_loss\) and the KL divergence term averaged over all samples in the batch:
%%   \[ \text{{vae\_loss}} = \text{{mean}}(\text{{xent\_loss}} + \text{{kl\_loss}}) \]
\begin{equation}
\text{{vae\_loss}} = \text{{mean}}(\text{{xent\_loss}} + \text{{kl\_loss}})  \label{vae_loss}
\end{equation}

The objective during training is to minimize this combined loss, leading to a VAE model that effectively reconstructs input data while regularizing the distribution of latent variables. We trained the autoencoder with 50 epochs and 128 batch size. The encoder part of the trained autoencoder is then isolated and used to transform the dataset into a lower-dimensional space, yielding the encoded features. This process involves passing the input data through the encoder model, where each sample is mapped to a reduced-dimensional representation. The resulting data contain the encoded features, which capture essential patterns and representations in the original data. These encoded features can subsequently be utilized for various downstream tasks, such as classification, offering a more compact and informative representation of the input data in a lower-dimensional feature space. After several trials and investigations of the parameters and features set, we choose 100 features from the 182 original features as our output set. This reduced feature set can be used for further analysis or as input features for other models. The selection is based on how much each feature contributes to the encoding layer, capturing the most relevant information according to the learned representation in the autoencoder. Figure \ref{fig:system} part one shows the  VAE architecture.

\subsubsection{Self-training}
Self-training is a semi-supervised learning where a model is initially trained on a small labeled dataset and then iteratively improves itself by incorporating unlabeled data. Self-training in semi-supervised learning is particularly useful when obtaining labeled data, which is expensive or time-consuming, as it allows leveraging a combination of labeled and unlabeled data to improve model performance. The process begins with training a model on the labeled dataset. This model then generates predictions for the unlabeled data, which are treated as pseudo-labels. These pseudo-labeled samples are added to the dataset, creating an augmented training set that combines original labeled data with newly generated pseudo-labeled examples. The model is re-trained on this expanded dataset, gradually improving its generalization ability. This essentially creates a larger dataset with a mix of labeled and pseudo-labeled examples~\cite{van2020survey}.

\begin{figure}[h!]
 \centering
  %\vspace*{-.45in}
 \subfloat{\includegraphics[width=16cm,height=5cm]{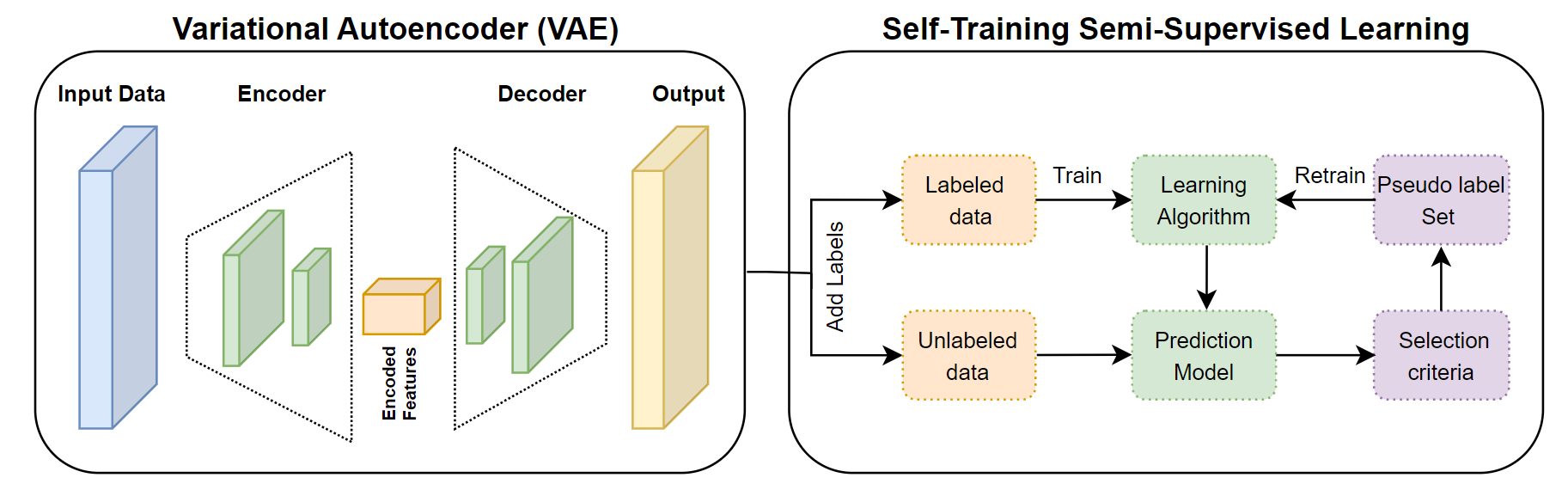}}
 \caption{Proposed System Architecture to Classify AA in PwD using VAE and Self-training.}
  \label{fig:system}
\end{figure}

The process is repeated, with the updated model making predictions on additional unlabeled data, generating new pseudo-labels, and adding them to the training set. However, it is crucial to manage the quality of pseudo-labels, as incorrect pseudo-labels can mislead the model during training. Techniques such as thresholding or using a confidence score can be employed to filter out unreliable pseudo-labels. This work sets the decision threshold for adding pseudo-labels to 0.7. If the predicted probability of a sample being in a particular class is above this threshold, it will be added to the training set with the pseudo-label. We also set the maximum number of iterations allowed for the self-training process to 100. Figure \ref{fig:system} part two shows the Self-training semi-supervised learning process.

\subsection{Performance Evaluation Metrics}
We use three classification models: Extra Trees, Random Forest, and XGBoost classification models to classify the Self-training model-generated labels. Extra Trees and Random Forest are bagging ensemble methods that create multiple decision trees and aggregate predictions to avoid overfitting. Extra Trees introduces extreme randomness in feature selection. In contrast, XGBoost employs a boosting strategy, sequentially building trees to correct errors and optimizing performance through gradient descent and regularization. These models proved to be the most promising models to classify AA in PwD in our previous work~\cite{10288764,10371835}. We used a random state for the three models to ensure random splitting is the same every time you run the code. It helps in obtaining consistent results during different runs. We also used stratified sampling to ensure that the distribution of the target variable y is similar in both the training and testing sets. This is particularly useful for imbalanced datasets such as our dataset to maintain the same proportion of different classes in both sets. We then split the dataset into training and testing sets, with 70\% used for training and 30\% for testing. 

On the other hand, we address the imbalanced data problem by selecting a performance evaluation technique that focuses on balanced performance evaluation to overcome this problem. Imbalanced datasets refer to scenarios where the distribution of classes is uneven, with one class significantly outnumbering the other(s). Traditional metrics like accuracy may not comprehensively understand a model's performance in such cases. We extend the conventional classification report by incorporating additional metrics specifically designed to address the challenges posed by imbalanced datasets. Accurate performance evaluation techniques played a pivotal role by providing an intricate breakdown of key metrics when dealing with imbalanced data. We use the balanced accuracy score function, which offers a balance between sensitivity and specificity. This metric accounts for class imbalance, providing a fair classification accuracy assessment. We also use Precision, which represents the proportion of true positive predictions among all positive predictions and showcases the accuracy of positive predictions. Conversely, recall signifies the proportion of true positive predictions among all actual positive instances, offering insights into the model's ability to capture all relevant cases.

The F1-score, the mean of Precision and recall, provides a balanced measure of a model's overall performance. Furthermore, incorporating the Area Under the Receiver Operating Characteristic Curve (AUC) allowed for a nuanced examination of the model's discriminatory power. By weighing the true positive rate against the false positive rate, AUC provided a robust measure of the model's ability to distinguish between classes. In addition to these traditional metrics, the evaluation extended to processing time as a practical consideration. Processing time measures the time required for the classification system to make predictions on the dataset. This aspect is crucial for real-world applications where efficiency is a paramount concern. Thus, the holistic evaluation framework considered the standard metrics like Precision, recall, and F1-score and incorporated processing time, providing a comprehensive understanding of the classification system's performance in the context of PwD and imbalanced datasets.

\section{Results and Discussion}\label{s:numerical}
In this section, we present the results of our experiments evaluating our proposed system, which utilizes VAE and Self-training, in comparison with different approaches for AA detection in PwD. To evaluate the proposed system, we compare the system with a fully supervised learning approach. For supervised learning, we only use the labeled data from our dataset. We had a total of 18804 minutes labeled normal and 1475 minutes labeled AA. We start by presenting the results from the baseline, which uses extracted features only with fully labeled data (5 participants) employing supervised learning models. Then, we introduce the proposed VAE for feature representation, which we integrate on top of the features extracted with the fully supervised model. Next, we present the proposed self-training model with extracted features, which includes the full dataset with partially labeled data (14 participants). Lastly, we present our proposed system, which utilizes the VAE features with the self-training model that incorporates the full dataset with partially labeled data (14 participants). Our aim is to compare all scenarios with our proposed system to demonstrate performance across different parameters and approaches. For the discussion, we compare our work with previous works on agitation detection and wearable sensors.

\subsection{Agitation Detection using Supervised Learning}

Table \ref{tab:table2} presents the baseline model with supervised learning and feature extraction using three prominent classification algorithms—Random Forest, Extra Trees, and XGBoost—to classify Normal and AA activities in PwD. In the supervised learning scenario, feature engineering techniques were applied to a dataset consisting of 18,804 minutes labeled as normal and 1,475 minutes labeled as AA from 5 participants. The input dataset comprised 182 features extracted from heart rate, accelerometer, electrodermal activity (EDA), and temperature measurements obtained from the E4 wristband data. Random Forest achieved a balanced accuracy of 78.50\%, with precision, recall, and F1-score of 96.9\%, 96.8\%, and 96.5\%, respectively. Extra Trees exhibited improved performance with a balanced accuracy of 79.01\%, precision, recall, and F1-score of 97.0\%, 96.9\%, and 96.6\%, respectively. XGBoost outperformed the other models, achieving a remarkably balanced accuracy of 85.03\%, with precision, recall, and F1-score of 97.8\%, 97.8\%, and 97.6\%, respectively. Random Forest, Extra Trees, and XGBoost demonstrated AUC ROC values of 98.74\%, 98.1\%, and 98.5\%, respectively. 

\begin{table}[h]
\centering
\caption{Baseline model with Fully Supervised Learning and Features}
\label{tab:table2}
\small % Adjusts font size to match the main document
\begin{tabular}{c|ccc}
\hline
Evaluation Metric &
  \multicolumn{1}{c}{Random Forest} &
  \multicolumn{1}{c}{Extra Trees} &
  XGBoost \\ 
 \hline
 Normal/AA Labels &
  \multicolumn{1}{c}{18804/1475} &
  \multicolumn{1}{c}{18804/1475} &
  18804/1475 \\
  \hline
  Balanced Accuracy &
  \multicolumn{1}{c}{78.50\%} &
  \multicolumn{1}{c}{79.01\%} &
  \textbf{85.03\%} \\
  \hline
  Precision &
  \multicolumn{1}{c}{96.9\%} &
  \multicolumn{1}{c}{97.0\%} &
  \textbf{97.8\%} \\
  \hline
  Recall &
  \multicolumn{1}{c}{96.8\%} &
  \multicolumn{1}{c}{96.9\%} &
  \textbf{97.8\%} \\
  \hline
  F1-Score &
  \multicolumn{1}{c}{96.5\%} &
  \multicolumn{1}{c}{96.6\%} &
  \textbf{97.6\%} \\
  \hline
  AUC ROC  &
  \multicolumn{1}{c}{98.7\%} &
  \multicolumn{1}{c}{98.1\%} &
  \textbf{98.5\%} \\
  \hline
  Processing Time &
  \multicolumn{1}{c}{19.3 s} &
  \multicolumn{1}{c}{\textbf{3.6 s}} &
  12 s \\
\hline
\end{tabular}
\end{table}

\subsection{Agitation Detection using Supervised Learning and VAE}
This section presents the supervised learning approach for classifying AA in PwD after using THE VAE. We trained the VAE model for 50 epochs. The output indicates that after 50 epochs, the loss on both the training and validation has decreased. The reported values of 0.0019 loss for training and 0.0017 loss for validation represent the mean squared error loss, measuring the dissimilarity between the reconstructed output and the input data. Lower loss values suggest that the autoencoder is successfully learning to reconstruct the input data, capturing essential features in the process. The output of the VAE was used as the input of the supervised models, which have 100 data features. Our analysis involves three prominent classification algorithms: Random Forest, Extra Trees, and XGBoost, utilizing the dataset with 100 features. For the supervised learning scenario shown in Table \ref{tab:table4}, the model is trained and evaluated using labeled data, consisting of 18,804 minutes labeled as normal and 1,475 minutes labeled as AA. The Balanced Accuracy values for Random Forest, Extra Trees, and XGBoost were 83.1\%, 81.5\%, and 86.4\%, respectively. Precision, recall, and F1-Score metrics revealed that Random Forest achieved 96.3\%, 96.1\%, and 95.4\%, Extra Trees reached 97.4\%, 97.3\%, and 97.1\%, and XGBoost attained 97.9\%, 97.9\%, and 97.7\%. AUC ROC scores were 98.0\% for Random Forest, 98.7\% for Extra Trees, and 98.8\% for XGBoost as shown in Table \ref{tab:table4}. 

\begin{table}[h!]
\centering
\caption{Baseline model with Fully Supervised Learning and Variational Autoencoder}
\label{tab:table4}
\small % Adjusts the font size to match the document
\begin{tabular}{c|ccc}
\hline
Evaluation Metric &
  \multicolumn{1}{c}{Random Forest} &
  \multicolumn{1}{c}{Extra Trees} &
  XGBoost \\ 
 \hline
Normal/AA Labels & \multicolumn{1}{c}{55214/2523} &
  \multicolumn{1}{c}{55648/2092} &
  54768/2974 \\ 
\hline
  Balanced Accuracy &
    \multicolumn{1}{c}{83.1\%} &
  \multicolumn{1}{c}{81.5\%} &
  \textbf{86.4\%} \\
\hline
  Precision &
  \multicolumn{1}{c}{96.3\%} &
  \multicolumn{1}{c}{97.4\%} &
  \textbf{97.9\%} \\
\hline
  Recall &
  \multicolumn{1}{c}{96.1\%} &
  \multicolumn{1}{c}{97.3\%} &
  \textbf{97.9\%} \\
\hline
  F1-Score &
   \multicolumn{1}{c}{95.4\%} &
  \multicolumn{1}{c}{97.1\%} &
  \textbf{97.7\%} \\
\hline
  AUC ROC &
  \multicolumn{1}{c}{98\%} &
  \multicolumn{1}{c}{98.7\%} &
  \textbf{98.8\%} \\
\hline
  Processing Time &
 \multicolumn{1}{c}{18 s} &
  \multicolumn{1}{c}{\textbf{4.4 s}} &
  9.7 s \\
\hline
\end{tabular}
\end{table}

\subsection{Agitation Detection using Self-training}

In the semi-supervised learning approach, the setting extends the dataset to include 37463 unlabelled data, 18804 labeled as normal, and 1,475 minutes labeled as AA. The input dataset comprised 182 features extracted from heart rate, accelerometer, EDA, and temperature measurements. We train the model using Self-training for our three classification techniques. After training, the dataset was extended to include 55,214/2,523 minutes Normal/AA for Random Forest, 55,648/2,092 minutes Normal/AA for Extra Trees, and 54,768/2,974 minutes Normal/AA for XGBoost as shown in Table \ref{tab:table3}. Self-training was employed as a semi-supervised learning strategy, utilizing the unlabeled dataset to further enhance model performance. Random Forest in the semi-supervised setting achieved a balanced accuracy of 82.78\%, with precision, recall, and F1-score of 97.9\%, 97.9\%, and 98.6\%, respectively with ten iterations. 

Extra Trees exhibited a balanced accuracy of 81.0\%, precision, recall, and F1-score of 97.9\%, 98.0\%, and 97.7\%, respectively with three iterations. XGBoost excelled with a balanced accuracy of 86.02\%, precision, recall, and F1-score of 98.4\%, 98.4\%, and 98.3\%, respectively with seven iterations. Random Forest, Extra Trees, and XGBoost achieved AUC ROC values of 99.01\%, 99.0\%, and 99.4\%, respectively. These detailed results highlight the effectiveness of both supervised and semi-supervised approaches, offering insights into the classification performance and processing efficiency of different models in the context of dementia-related activity classification. The utilization of self-training in semi-supervised learning further enhances the models' ability to learn from unlabeled data, contributing to improved performance metrics.

\begin{table}[h]
\centering
\caption{Self-training with Features}
\label{tab:table3}
\small % Adjusts font size to match document style
\begin{tabular}{c|ccc}
\hline
Evaluation Metric &
  \multicolumn{1}{c}{Random Forest} &
  \multicolumn{1}{c}{Extra Trees} &
  XGBoost \\ 
 \hline
Normal/AA Labels & \multicolumn{1}{c}{55214/2523} &
  \multicolumn{1}{c}{55648/2092} &
  54768/2974 \\ 
\hline
  Balanced Accuracy &
   \multicolumn{1}{c}{82.78\%} &
  \multicolumn{1}{c}{81.0\%} &
  \textbf{86.02\%}  \\
\hline
  Precision &
   \multicolumn{1}{c}{97.9\%} &
  \multicolumn{1}{c}{97.9\%} &
  \textbf{98.4\%}  \\
\hline
  Recall &
  \multicolumn{1}{c}{97.9\%} &
  \multicolumn{1}{c}{98.0\%} &
  \textbf{98.4\%} \\
\hline
  F1-Score &
  \multicolumn{1}{c}{98.6\%} &
  \multicolumn{1}{c}{97.7\%} &
  \textbf{98.3\%} \\
\hline
  AUC ROC &
  \multicolumn{1}{c}{99\%} &
  \multicolumn{1}{c}{99\%} &
  \textbf{99.4\%} \\
\hline
  Processing Time &
 \multicolumn{1}{c}{25 min 30 s} &
  \multicolumn{1}{c}{\textbf{5 min 50 s}} &
  17 min 33 s \\
\hline
\end{tabular}
\end{table}

%\subsection{Proposed System: Agitation Detection using Self-training and VAE} \label{z}
\subsection{Agitation Detection using Self-training and VAE} \label{z}
In the semi-supervised learning approach with VAE as shown in Table \ref{tab:table5}, the dataset is augmented with additional unlabeled data, and a VAE is introduced for feature extraction, reducing the dimensionality to 100 from 182. The dataset includes 37463 unlabelled data, 18804 labeled as normal, and 1,475 minutes labeled as AA. Self-training is employed as a semi-supervised learning technique. The label increased after the Self-training for the Random Forest model to 54998 labeled normal and 2733 labeled AA, Extra Trees classifier to 55665 labeled normal and 2076 labeled AA, and XGBoost classifier to 54712 labeled normal and 3026 labeled AA.

\begin{table}[h!]
\centering
\caption{Self-training with Variational Autoencoder}
\label{tab:table5}
\small % Adjusts font size to maintain consistency with the document
\begin{tabular}{c|ccc}
\hline
Evaluation Metric &
  \multicolumn{1}{c}{Random Forest} &
  \multicolumn{1}{c}{Extra Trees} &
  XGBoost \\ 
\hline
Normal/AA Labels & \multicolumn{1}{c}{54998/2733} &
  \multicolumn{1}{c}{55665/2076} &
  54712/3026 \\ 
\hline
  Balanced Accuracy &
    \multicolumn{1}{c}{85.52\%} &
  \multicolumn{1}{c}{86.94\%} &
  \textbf{90.18\%}  \\
\hline
  Precision &
\multicolumn{1}{c}{98.0\%} &
  \multicolumn{1}{c}{98.90\%} &
  \textbf{99.0\%}  \\
\hline
  Recall &
  \multicolumn{1}{c}{98.0\%} &
  \multicolumn{1}{c}{98.90\%} &
  \textbf{99.0\%}  \\
\hline
  F1-Score &
   \multicolumn{1}{c}{98.0\%} &
  \multicolumn{1}{c}{98.90\%} &
  \textbf{99.0\%} \\
\hline
  AUC ROC &
  \multicolumn{1}{c}{98.3\%} &
  \multicolumn{1}{c}{99.5\%} &
  \textbf{99.6\%} \\
\hline
  Processing Time &
 \multicolumn{1}{c}{14 min 46 s} &
  \multicolumn{1}{c}{\textbf{3 min 38 s}} &
  13 min 30 s \\
\hline
\end{tabular}
\end{table}

\begin{figure}%
    \centering
    \subfloat[\centering AUC ROC Curve]{{\includegraphics[width=7cm]{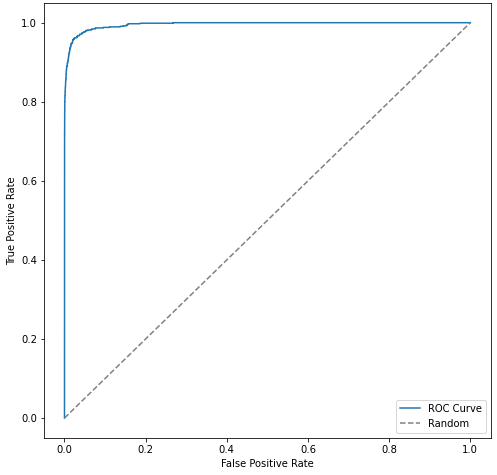} }}%
    \qquad
    \subfloat[\centering Precision-Recall Curve]{{\includegraphics[width=7cm]{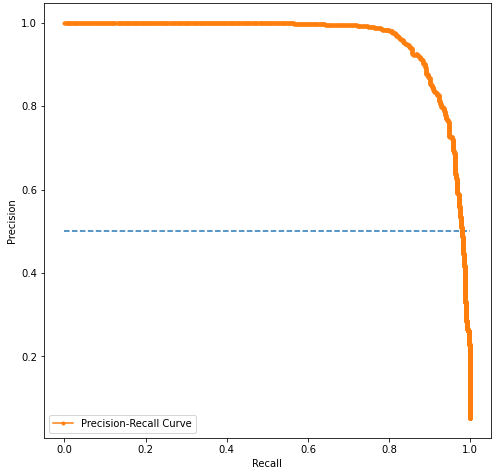} }}%
    \caption{AUC ROC and Precision-Recall Curve for XGBoost Classifier using VAE and Semi-Supervised Learning.}
    \label{fig:example}%
\end{figure}

 The balanced Accuracy for Random Forest, Extra Trees, and XGBoost reached 85.52\%, 86.94\%, and 90.18\%, respectively. Precision, Recall, and F1-Score values were 98.0\%, 98.0\%, and 98.0\% for Random Forest, 98.90\%, 98.90\%, and 98.90\% for Extra Trees, and 99.0\%, 99.0\%, and 99.0\% for XGBoost. AUC ROC scores stood at 98.3\% for Random Forest, 99.5\% for Extra Trees, and 99.6\% for XGBoost model. Figure \ref{fig:example} shows the AUC ROC and Precision-Recall Curve for XGBoost Classifier using VAE and Semi-Supervised Learning. The results showcase the efficacy of the semi-supervised learning approach, particularly when leveraging VAE-based feature extraction. The proposed system utilizing Self-training with VAE presents the best overall performance compared to fully supervised learning with features, fully supervised learning with VAE, and semi-supervised learning with features.

\subsection{Discussion}\label{s:discussion}
The results of our proposed system in subsection \ref{z} showcase promising outcomes for classifying AA in PwD with a partially labeled dataset when utilizing VAE and Self-training. We compare our proposed system with different approaches to present the performance in comparison with different methods. Leveraging self-training and VAE-based feature representation boosted the models'  overall performance. XGBoost, in particular, demonstrated remarkable results with a balanced accuracy of 90.18\% and precision, recall, and F1-score all reaching 99\%, indicating the effectiveness of VAE-based approaches. While XGBoost consistently outperformed other models, Extra Trees demonstrated the lowest processing time in both supervised and semi-supervised settings. This suggests a trade-off between processing efficiency and classification performance, with Extra Trees being a more time-efficient option. 

Incorporating unlabeled data through self-training significantly improves the models' ability to classify AA, as evidenced by higher balanced accuracy, precision, recall, and F1-score. The VAE-based approach, emphasizing learning a compact representation of features, further amplifies the models' performance. The reduction in feature dimensionality, coupled with the utilization of self-training, resulted in superior classification capabilities. The enhanced performance observed in semi-supervised learning, especially when incorporating VAE-based feature extraction, highlights the practical applicability of leveraging unlabeled data. In real-world scenarios where labeled data is often scarce, these approaches offer a promising avenue for improving the robustness of AA detection in PwD.

To evaluate our work in comparison with previous studies, we conducted a comprehensive search for papers addressing the detection of AA in PwD using the E4 wristband shown in Table \ref{tab:table_comparison}. Despite the fact that we tested our methodology on a distinct dataset and proposed system, which may render direct comparisons less equitable, we aim to present the conceptual framework of our work alongside that of others. In a study by Khan et al. ~\cite{spasojevic2021pilot}, the authors proposed a system for classifying AA in PwD, involving 17 participants with normal and AA labels and utilizing the E4 wristband. They extracted 67 time and frequency domain features, processing 82,364 samples from wristband data with a 1-minute window size. Using the Random Forest Classifier, their approach yielded an Area Under the Curve (AUC) Receiver Operating Characteristic (ROC) of 87\%. 

In one of our previous works, we re-examined the dataset used by Khan et al. ~\cite{spasojevic2021pilot}, employing the same feature engineering techniques outlined in this paper~\cite{10288764}. However, our work focused on investigating new sets of features using various machine learning models to discern personalized AA patterns. Notably, the Extra Trees model demonstrated superior performance, achieving a median AUC of 94.1\% across the combined patient average score. This proposed system collected data from 14 participants (5 fully labeled and 9 unlabeled), extracting 100 features from the time, frequency, and time-frequency domains after using Variational Autoencoders. Employing Semi-Supervised Learning, we labeled all data, comprising a total of 57,738 samples with a 1-minute window size. Notably, our approach achieved an impressive 99.6\% AUC ROC, outperforming prior works in this domain.
\begin{table}[h!]
\centering
\caption{Comparison of AUC ROC Values across Different Studies}
\label{tab:table_comparison}
\small % Adjusts font size to maintain consistency with the document
\begin{tabular}{c|ccc}
\hline
Evaluation Metric       & Khan et al.            & Our Previous Work        & Proposed System \\ 
\hline
Model Used              & Random Forest          & Extra Trees              & \makecell{Self-training\\ with XGBoost} \\ 
\hline
\# Participants         & 17                     & 17                       & 14             \\ 
\hline
\# Processed Samples    & 82,364                 & 82,364                   & 57,738         \\ 
\hline
\# Features             & 67                     & 60-120                   & 100            \\ 
\hline
AUC ROC                 & 87\%                   & 94.1\%                   & \textbf{99.6\%} \\ 
\hline
\end{tabular}
\end{table}

\section{Conclusion}\label{s:conclusion}
In conclusion, the proposed study provides a significant advancement in the area of partially labeled data using semi-supervised AA detection in PwD. It specifically incorporates AI with wearable sensor data that collects physiological biomarkers and directly addresses the issue of limited labeled real-world datasets. The datasets used in this study are composed of digital biomarkers collected by wearable sensors worn by patients from different sites across Canada.  Our solution introduces a novel application of semi-supervised learning combined with Variational Autoencoders (VAE) for feature extraction and selection. The comparison between the proposed systems and other methods underscores the effectiveness of VAE in improving classification metrics for AA in PwD. The results show that this combination of VAE for feature representation and Self-training for generating pseudo labels for the unlabeled data outperforms the baseline model with feature extraction only or VAE. Out of the three different classifiers used in the comparison between supervised and semi-supervised learning, XGBoost achieved the highest accuracy of 90.18\% in the case of semi-supervised learning coupled with VAE. The ability to learn and achieve better performance from limited labeled data showcases the potential for real-world applications in AA detection for PwD. The impact of such applications could transform patient care and improve the quality of life for the patients and those surrounding them. Moreover, these realistic applications that bridge the gap between AI enhancements and data availability open the path for further exploration and utilization of many limited labels gathered from healthcare.

\bibliographystyle{IEEEtran}
\bibliography{references.bib}

\begin{thebibliography}{10}
\providecommand{\url}[1]{#1}
\csname url@samestyle\endcsname
\providecommand{\newblock}{\relax}
\providecommand{\bibinfo}[2]{#2}
\providecommand{\BIBentrySTDinterwordspacing}{\spaceskip=0pt\relax}
\providecommand{\BIBentryALTinterwordstretchfactor}{4}
\providecommand{\BIBentryALTinterwordspacing}{\spaceskip=\fontdimen2\font plus
\BIBentryALTinterwordstretchfactor\fontdimen3\font minus \fontdimen4\font\relax}
\providecommand{\BIBforeignlanguage}[2]{{%
\expandafter\ifx\csname l@#1\endcsname\relax
\typeout{** WARNING: IEEEtran.bst: No hyphenation pattern has been}%
\typeout{** loaded for the language `#1'. Using the pattern for}%
\typeout{** the default language instead.}%
\else
\language=\csname l@#1\endcsname
\fi
#2}}
\providecommand{\BIBdecl}{\relax}
\BIBdecl

\bibitem{WHO2023}
W.~H. Organization, ``Dementia,'' \url{https://www.who.int/news-room/fact-sheets/detail/dementia}, 2023, [Accessed \today].

\bibitem{gale2018dementia}
S.~A. Gale, D.~Acar, and K.~R. Daffner, ``Dementia,'' \emph{Am J Med}, vol. 131, no.~10, pp. 1161--1169, 2018.

\bibitem{bansal2014dementia}
N.~Bansal and M.~Parle, ``Dementia: An overview,'' \emph{Journal of Pharmaceutical Technology, Research and Management}, vol.~2, pp. 29--45, 2014.

\bibitem{kester2009dementia}
M.~I. Kester and P.~Scheltens, ``Dementia: the bare essentials,'' \emph{Practical Neurology}, vol.~9, no.~4, pp. 241--251, 2009.

\bibitem{Lyketsos2002}
C.~G. Lyketsos, O.~Lopez, B.~Jones, A.~L. Fitzpatrick, J.~Breitner, and S.~DeKosky, ``Prevalence of neuropsychiatric symptoms in dementia and mild cognitive impairment: results from the cardiovascular health study,'' \emph{Jama}, vol. 288, no.~12, pp. 1475--1483, Sep 25 2002, (in eng).

\bibitem{burhan2023s13}
A.~M. Burhan, W.~Sun, M.~Chiu, S.~Choudhury, A.~Badawi, and K.~Elgazzar, ``S13: Technology enabled care for neuropsychiatric symptoms of dementia: implementation at the point of care,'' \emph{International Psychogeriatrics}, vol.~35, no.~S1, pp. 40--40, 2023.

\bibitem{Ballard2013}
C.~Ballard and A.~Corbett, ``Agitation and aggression in people with alzheimer's disease,'' \emph{Curr Opin Psychiatry}, vol.~26, no.~3, pp. 252--259, May 2013, (in eng).

\bibitem{khan2018detecting}
S.~S. Khan, B.~Ye, B.~Taati, and A.~Mihailidis, ``Detecting agitation and aggression in people with dementia using sensors—a systematic review,'' \emph{Alzheimer's \& Dementia}, vol.~14, no.~6, pp. 824--832, 2018.

\bibitem{cohen1990dementia}
J.~Cohen-Mansfield, M.~S. Marx, and A.~S. Rosenthal, ``Dementia and agitation in nursing home residents: How are they related?'' \emph{Psychology and Aging}, vol.~5, no.~1, p.~3, 1990.

\bibitem{bharucha2009intelligent}
A.~J. Bharucha, V.~Anand, J.~Forlizzi, M.~A. Dew, C.~F. Reynolds~III, S.~Stevens, and H.~Wactlar, ``Intelligent assistive technology applications to dementia care: current capabilities, limitations, and future challenges,'' \emph{The American journal of geriatric psychiatry}, vol.~17, no.~2, pp. 88--104, 2009.

\bibitem{fabrizio2021artificial}
C.~Fabrizio, A.~Termine, C.~Caltagirone, and G.~Sancesario, ``Artificial intelligence for alzheimer’s disease: promise or challenge?'' \emph{Diagnostics}, vol.~11, no.~8, p. 1473, 2021.

\bibitem{rezvani2021semi}
R.~Rezvani \emph{et~al.}, ``Semi-supervised learning for identifying the likelihood of agitation in people with dementia,'' \emph{Journal of Healthcare Informatics Research}, vol.~5, pp. 45--58, 2021.

\bibitem{Sato2019AutomaticFE}
K.~Sato, M.~Chida, Y.~Hayakawa, and N.~M. Fujiki, ``Automatic feature extraction from wearable sensor data by use of machine learnings,'' \emph{Proceedings of The 7th International Conference on Intelligent Systems and Image Processing 2019}, 2019.

\bibitem{Chikhaoui2017TowardsAF}
B.~Chikhaoui and F.~Gouineau, ``Towards automatic feature extraction for activity recognition from wearable sensors: A deep learning approach,'' \emph{2017 IEEE International Conference on Data Mining Workshops (ICDMW)}, pp. 693--702, 2017.

\bibitem{abdallah2018activity}
Z.~S. Abdallah, M.~M. Gaber, B.~Srinivasan, and S.~Krishnaswamy, ``Activity recognition with evolving data streams: A review,'' \emph{ACM Computing Surveys (CSUR)}, vol.~51, no.~4, p.~71, 2018.

\bibitem{Li2020ExtractionAI}
B.~Li and A.~Sano, ``Extraction and interpretation of deep autoencoder-based temporal features from wearables for forecasting personalized mood, health, and stress,'' \emph{Proceedings of the ACM on Interactive, Mobile, Wearable and Ubiquitous Technologies}, vol.~4, pp. 1 -- 26, 2020.

\bibitem{kingma2013auto}
D.~P. Kingma and M.~Welling, ``Auto-encoding variational bayes,'' \emph{arXiv preprint arXiv:1312.6114}, 2013.

\bibitem{NCT04516057}
\BIBentryALTinterwordspacing
``Nabilone for agitation blinded intervention trial (nab-it),'' ClinicalTrials.gov, 2021, [Accessed \today]. [Online]. Available: \url{https://clinicaltrials.gov/ct2/show/NCT04516057}
\BIBentrySTDinterwordspacing

\bibitem{NCT03672201}
\BIBentryALTinterwordspacing
``Standardizing care for neuropsychiatric symptoms and quality of life in dementia (stan),'' ClinicalTrials.gov, [Accessed \today]. [Online]. Available: \url{https://clinicaltrials.gov/ct2/show/NCT03672201}
\BIBentrySTDinterwordspacing

\bibitem{spasojevic2021pilot}
S.~Spasojevic, J.~Nogas, A.~Iaboni, B.~Ye, A.~Mihailidis, A.~Wang, S.~J. Li, L.~S. Martin, K.~Newman, and S.~S. Khan, ``A pilot study to detect agitation in people living with dementia using multi-modal sensors,'' \emph{Journal of Healthcare Informatics Research}, vol.~5, no.~3, pp. 342--358, 2021.

\bibitem{nesbitt201915}
C.~Nesbitt, A.~Gupta, K.~Maly, H.~R. Okhravi, and S.~Jain, ``15 feasibility of using wearable sensors to detect agitation in persons with dementia,'' \emph{CNS Spectrums}, vol.~24, no.~1, pp. 181--181, 2019.

\bibitem{khan2019agitation}
S.~S. Khan, S.~Spasojevic, J.~Nogas, B.~Ye, A.~Mihailidis, A.~Iaboni, A.~Wang, L.~S. Martin, and K.~Newman, ``Agitation detection in people living with dementia using multimodal sensors,'' in \emph{2019 41st Annual International Conference of the IEEE Engineering in Medicine and Biology Society (EMBC)}.\hskip 1em plus 0.5em minus 0.4em\relax IEEE, 2019, pp. 3588--3591.

\bibitem{zhang2022deep}
S.~Zhang, Y.~Li, S.~Zhang, F.~Shahabi, S.~Xia, Y.~Deng, and N.~Alshurafa, ``Deep learning in human activity recognition with wearable sensors: A review on advances,'' \emph{Sensors}, vol.~22, no.~4, p. 1476, 2022.

\bibitem{chikhaoui2017towards}
B.~Chikhaoui and F.~Gouineau, ``Towards automatic feature extraction for activity recognition from wearable sensors: a deep learning approach,'' in \emph{2017 IEEE international conference on data mining workshops (ICDMW)}.\hskip 1em plus 0.5em minus 0.4em\relax IEEE, 2017, pp. 693--702.

\bibitem{wang2016recognition}
L.~Wang, ``Recognition of human activities using continuous autoencoders with wearable sensors,'' \emph{Sensors}, vol.~16, no.~2, p. 189, 2016.

\bibitem{zhu2022introduction}
X.~Zhu and A.~B. Goldberg, \emph{Introduction to semi-supervised learning}.\hskip 1em plus 0.5em minus 0.4em\relax Springer Nature, 2022.

\bibitem{hassanzadeh2018clinical}
H.~Hassanzadeh, M.~Kholghi, A.~Nguyen, and K.~Chu, ``Clinical document classification using labeled and unlabeled data across hospitals,'' in \emph{AMIA annual symposium proceedings}, vol. 2018.\hskip 1em plus 0.5em minus 0.4em\relax American Medical Informatics Association, 2018, p. 545.

\bibitem{tarvainen2017mean}
A.~Tarvainen and H.~Valpola, ``Mean teachers are better role models: Weight-averaged consistency targets improve semi-supervised deep learning results,'' in \emph{Advances in neural information processing systems}, vol.~30, 2017, pp. 1195--1204.

\bibitem{triguero2015selflabeled}
I.~Triguero, S.~Garc{\'i}a, and F.~Herrera, ``Self-labeled techniques for semi-supervised learning: Taxonomy, software and empirical study,'' \emph{Knowledge and Information Systems}, vol.~42, no.~2, pp. 245--284, 2015.

\bibitem{sarkar2020selfsupervised}
P.~Sarkar and A.~Etemad, ``Self-supervised ecg representation learning for emotion recognition,'' \emph{arXiv preprint arXiv:2002.03898}, 2020.

\bibitem{quispe2021applying}
K.~G.~M. Quispe \emph{et~al.}, ``Applying self-supervised representation learning for emotion recognition using physiological signals,'' \emph{IEEE Transactions on Affective Computing}, vol.~9, pp. 78--91, 2021.

\bibitem{doe2021semi}
J.~Doe \emph{et~al.}, ``Semi-supervised learning for emotion recognition in dementia care: A review,'' \emph{International Journal of Dementia Research}, vol.~7, pp. 65--80, 2021.

\bibitem{hinkle2021endtoend}
L.~B. Hinkle \emph{et~al.}, ``An end-to-end methodology for semi-supervised har data collection, labeling, and classification using a wristband,'' \emph{Journal of Ambient Intelligence and Humanized Computing}, vol.~11, pp. 215--230, 2021.

\bibitem{yu2022semi}
H.~Yu and A.~Sano, ``Semi-supervised learning and data augmentation for wearable-based health monitoring system in the wild,'' in \emph{NeurIPS 2022 Workshop on Learning from Time Series for Health}, 2022.

\bibitem{hekmatiathar2022data}
S.~HekmatiAthar \emph{et~al.}, ``Data-driven forecasting of agitation for persons with dementia: A deep learning-based approach,'' \emph{International Journal of Dementia Care}, vol.~8, pp. 112--126, 2022.

\bibitem{american2013diagnostic}
A.~P. Association, \emph{Diagnostic and Statistical Manual of Mental Disorders}, 5th~ed.\hskip 1em plus 0.5em minus 0.4em\relax Arlington, VA, US: American Psychiatric Publishing, Inc., 2013.

\bibitem{Cummings2015AgitationCognitiveDisorders}
J.~Cummings \emph{et~al.}, ``Agitation in cognitive disorders: International psychogeriatric association provisional consensus clinical and research definition,'' \emph{International Psychogeriatrics}, vol.~27, no.~1, pp. 7--17, 2015.

\bibitem{kurlowicz1999mini}
L.~Kurlowicz and M.~Wallace, ``The mini-mental state examination (mmse),'' pp. 8--9, 1999.

\bibitem{cummings2015agitation}
J.~Cummings, J.~Mintzer, H.~Brodaty, M.~Sano, S.~Banerjee, D.~Devanand, S.~Gauthier, R.~Howard, K.~Lanct{\^o}t, C.~G. Lyketsos \emph{et~al.}, ``Agitation in cognitive disorders: International psychogeriatric association provisional consensus clinical and research definition,'' \emph{International psychogeriatrics}, vol.~27, no.~1, pp. 7--17, 2015.

\bibitem{empactica}
\BIBentryALTinterwordspacing
{Empatica Inc.}, ``{Empatica | Medical devices, AI and algorithms for remote patient monitoring},'' Empatica, [Accessed \today]. [Online]. Available: \url{https://www.empatica.com}
\BIBentrySTDinterwordspacing

\bibitem{10288764}
A.~Badawi, K.~Elgazzar, B.~Ye, K.~Newman, A.~Mihailidis, A.~Iaboni, and S.~S. Khan, ``Investigating multimodal sensor features importance to detect agitation in people with dementia,'' in \emph{2023 IEEE Canadian Conference on Electrical and Computer Engineering (CCECE)}, 2023, pp. 77--82.

\bibitem{10371835}
A.~Badawi, S.~Choudhury, K.~Elgazzar, and A.~M. Burhan, ``Artificial intelligence and features investigating to detect neuropsychiatric symptoms in patients with dementia: A pilot study,'' in \emph{2023 IEEE Symposium Series on Computational Intelligence (SSCI)}, 2023, pp. 741--746.

\bibitem{foll2021flirt}
S.~F{\"o}ll, M.~Maritsch, F.~Spinola, V.~Mishra, F.~Barata, T.~Kowatsch, E.~Fleisch, and F.~Wortmann, ``Flirt: A feature generation toolkit for wearable data,'' \emph{Computer Methods and Programs in Biomedicine}, vol. 212, p. 106461, 2021.

\bibitem{kingma2019introduction}
D.~P. Kingma and M.~Welling, ``An introduction to variational autoencoders,'' \emph{arXiv preprint arXiv:1906.02691}, 2019.

\bibitem{doersch2016tutorial}
C.~Doersch, ``Tutorial on variational autoencoders,'' \emph{arXiv preprint arXiv:1606.05908}, 2016.

\bibitem{van2020survey}
J.~E. Van~Engelen and H.~H. Hoos, ``A survey on semi-supervised learning,'' \emph{Machine learning}, vol. 109, no.~2, pp. 373--440, 2020.

\end{thebibliography}
\end{document}